\documentclass[preprint,review,12pt]{elsarticle}
\usepackage{amssymb, amsmath, amsthm, bm}
\usepackage[noend]{algpseudocode}
\usepackage{algorithmicx,algorithm}
\usepackage{graphicx}
\usepackage{caption}
\usepackage[colorlinks,linkcolor=blue]{hyperref}
\usepackage{epstopdf}
\usepackage[colorlinks,linkcolor=blue]{hyperref}
\usepackage{bbding}
\usepackage{booktabs}
\usepackage{tikz,xcolor}

\newtheorem{rmrk}{Remark}
\numberwithin{equation}{section}

\journal{}

\begin{document}
\begin{frontmatter}
\title{An Orthogonal Classifier for Improving the Adversarial Robustness of Neural Networks }
\author[label1] {Cong Xu \fnref{cor1}}
\author[label2] {Xiang Li \fnref{cor2}}
\author[label1] {Min Yang \corref{cor3}}
\fntext[cor1] {Email: congxueric@gmail.com }
\fntext[cor2] {Email: lixiang2020ecnu@163.com}
\cortext[cor3] {Corresponding author: yang@ytu.edu.cn}

\address[label1]{School of Mathematics and Information Sciences, Yantai University, Yantai 264005, China}
\address[label2]{Software Engineering Institute, East China Normal University, Shanghai, China}

	\begin{abstract}
		Neural networks are susceptible to artificially designed adversarial perturbations.
		Recent efforts have shown that imposing certain modifications on classification layer can improve the robustness of the neural networks.
        In this paper, we explicitly construct a dense orthogonal weight matrix whose entries have the same magnitude, thereby leading to a novel robust classifier.
	    The proposed classifier avoids the undesired structural redundancy issue in previous work.
        Applying this classifier in standard training on clean data is sufficient to ensure the high accuracy and good robustness of the model.
        Moreover,  when extra adversarial samples are used, better robustness can be further obtained with the help of a special worst-case loss.
        Experimental results  show that our method is efficient and competitive to many state-of-the-art defensive approaches.
        Our code is available at \url{https://github.com/MTandHJ/roboc}.
    \end{abstract}

	
\end{frontmatter}

	\section{Introduction}
	
	Neural networks have achieved impressive performance in many challenging computer vision tasks \cite{he2016,krizhevsky2009,lecun1998,wang2019}.
	But it was soon realized that they are vulnerable to artificial adversarial perturbations \cite{goodfellow2015,szegedy2013,wang2021,xu2020}.
	Imposing human imperceptible perturbations on clean data could deceive the networks and cause incorrect classification.
	In order to improve the robustness of neural networks,
    a large number of studies on defense mechanisms have appeared; see, e.g., \cite{cohen2019,gowal2019,guo2019,han2019,madry2018,wang2020,zhang2019b}.
    Although these methods have achieved good performance,
    the training process is usually computationally expensive and not scalable \cite{shafahi2019,zhang2019a}.

    It was pointed out by \cite{moosavi2016} that the feature representations learned by standard training tend to concentrate on decision boundaries,
	an undesirable property resulting in vulnerability of networks.
    By analyzing the relationship between robustness and feature distribution,
    some recent studies intend to use the modified classifiers to improve the robustness of the networks \cite{mustafa2019,han2019,pang2018,pang2020}.
	The corresponding approaches seem more scalable and easier to implement.

	In particular, starting from the Max-Mahalanobis distribution,
    Pang et al. \cite{pang2018,pang2020} proposed an algorithm called GenerateOptMeans to explicitly build a linear classifier,
 	which brings a significant improvement in adversarial robustness with very little additional computational cost.
    However, the classifier generated by GenerateOptMeans suffers from the problem of structural redundancy
    (see detailed discussions in Section \ref{sec:construction} and experimental results in Section \ref{experiment:redundancy}),
    which may lead to underfitting in some cases.

    In this article, we are to introduce a class of dense orthogonal vectors to construct a novel robust classifier.
    Compared to the GenerateOptMeans algorithm used in \cite{pang2018, pang2020},
    the classifier weights calculation procedure we give is much simpler and can avoid the undesired structural redundancy issue.
    In addition, orthogonal weights are also conducive to learning more discriminative features while ensuring a stable training \cite{wang2019}.
    Experimental results show that using the proposed classifier in standard training on clean data is sufficient to ensure the high accuracy and good robustness of the model.
    Moreover, if extra adversarial samples are available,
    the robustness can be further improved with the help of a worst-case intra-class compactness loss.
    Comparative empirical results on MNIST, FashionMNIST and CIFAR-10 datasets
    show that the proposed approach achieves the state-of-the-art robustness under various attacks,
    while still maintaining high accuracy on clean data.

	The main contributions of the work are summarized as follows:	
	\begin{itemize}
		\item [(1)]  We use a special type of dense orthogonal matrix,
                          whose entries have the same magnitude, to construct a novel classifier to improve the  robustness of neural networks.
                          The proposed classifier encourages the encoder to learn the feature representations that are compact within the class and dispersed between classes.

		\item [(2)]  We provide a step by step procedure to explicitly determine the weights of the classifier.
		                  Compared to the GenerateOptMeans algorithm developed in \cite{pang2018,pang2020},
		                  our procedure is much simpler and avoids the structural redundancy issue.

		\item [(3)]  Our classifier can be used in two ways.
                          One is only to perform standard training on clean data.
                          The other is to use extra adversarial samples and a corresponding worst-case  intra-class compactness loss for adversarial training.
					      The former keeps a good balance between the efficiency, accuracy and  robustness.
                          The latter guarantees a better robustness performance.
	\end{itemize}

	\section{Related Works}
	
	\subsection{Orthogonal neural networks}
	There has been a lot of work on orthogonal neural networks, see e.g. \cite{bansal2018,huang2018,wang2019,xie2017}.
	The use of orthogonal weights in neural networks can encourage the model to learn diverse and expressive features,
	and simultaneously improve the training stability \cite{bansal2018,wang2019}.
	A common method of introducing orthogonality into networks is to construct specific regularized terms in the optimization objective.

    In this paper, we aim to design an orthogonal classifier to improve the adversarial robustness of the networks.
    The main difference between our work and the previous research lies in two aspects.
    First, the previous research basically considered the orthogonality of the entire network,
    whereas this article only focuses on the classifier part, i.e., the penultimate layer.
    Second,  instead of using the common soft regularization approach,
    we present an explicit procedure to  construct the required orthogonal weights directly.
    And the orthogonal weights are frozen during the training.

	\subsection{Attack models and defense methods}
	Since the discovery that neural networks are vulnerable to artificial perturbations,
	a series of attack methods have emerged to evaluate the robustness of networks
	\cite{carlini2017,croce2020,moosavi2016,madry2018,foolbox2017,szegedy2013,tramer2019,wang2021}.
   These attack models design small perturbations to clean samples to generate various adversarial examples.
    FGSM  \cite{goodfellow2015} is a simple but effective attack method that utilizes the sign of the gradients of the loss function to generate adversarial samples.
	PGD \cite{madry2018}  is a more powerful iterative attack that starts from a random position
	in the neighborhood of a clean input and then applies FGSM for several iterations.
	DeepFool \cite{moosavi2016} is a gradient-based attack algorithm that iteratively linearizes the classifier
    to generate the smallest perturbation sufficient to change the classification label.
	C\&W \cite{carlini2017} is one of the most powerful attack to detect adversarial samples in $\ell_2$ norm.
	SLIDE \cite{tramer2019} is a more efficient model that  overcomes the inefficiency of PGD in searching for $\ell_1$ perturbations.
	Recently, Croce et al. \cite{croce2020} combined four diverse attacks into a more aggressive one, AutoAttack,
	as a new benchmark for empirical robustness evaluation.

	With the development of attack models,
    corresponding defense mechanisms have been continuously strengthened;
	see, e.g., adversarial training \cite{goodfellow2015,madry2018,wang2021,zhang2019b}, certified robustness \cite{cohen2019,gowal2019} and detection methods \cite{guo2019}.
    Among them,  adversarial training \cite{madry2018}, which minimizes the worst-case loss in the perturbation region, is considered to be one of the most powerful defenses.
	However, this defense is difficult to apply to large-scale problems due to its heavy computational cost.
	Another problem with adversarial training is that improved robustness often comes at the expense of the prediction accuracy of clean data \cite{su2018,zhang2019b}.
    Some recent work \cite{pang2018,pang2020,wan2018}  resorted to modify the linear classifier part to improve the robustness of the networks,
    without bringing too much extra computation.
	Particularly, Pang et al. \cite{pang2018,pang2020} provided a procedure called GenerateOptMeans algorithm
    to explicitly preset the weights of the classification layer,
    thereby helping improve the robustness of the trained model.
	However, the classifier derived in \cite{pang2018,pang2020} suffers from the problem of structural redundancy,
    which may lead to underfitting in some cases (see discussions in Section \ref{sec:construction}).

	\section{Orthogonal Classifier for Robustness Improvement }

	\subsection{The relationship between robustness and feature distribution}
	\label{sec:objectives}
	In this section, we first briefly discuss the relationship between robustness and feature distribution from an intuitive perspective,
	and then derive the corresponding optimization objectives.
	
	Let $ P $ denote the dimension of the feature representations, and $ K $ be the number of categories.
	A typical feed-forward neural network includes a nonlinear encoder $ f $ that
	transforms the input $x$ to its feature representation $f(x) \in \mathbb{R}^{P}$,
	and a multi-class classifier $ c $  given by
	\begin{align}
		c(x) = W^Tf(x) + b,
	\end{align}
	where $W \in \mathbb{R}^{P \times K}$ and $b \in \mathbb{R}^{K}$ denote the weight matrix and bias, respectively.

    Since the bias has little impact on the prediction,
    we will assume that $ b $ is zero for brevity  in the analysis below.
	For $ 1\leq i \leq K $, define the classification domain:
	\begin{align}
		\label{region}
		\mathcal{A}_i :=\{f \in \mathbb{R}^P: (w_i - w_j)^T f \ge 0, j \not = i \},
	\end{align}
	where $ w_i $ is the $ i $-th column of the weight matrix $W$.
	If the feature representation of a sample is located in domain $ A_i $,
    then we think the sample belongs to class $i $.

	Thus, for a given sample $ x $ with the ground-truth label $ y $,
    in order to correctly classify it,  the following estimate
	\begin{align}
		w_y^T f \geq  w_j^T f , \quad \forall j \not =y
	\end{align}
	should hold true. That is
	\begin{align}
		\label{normeq2}
		||w_y||_2 \cos \theta_y \geq  ||w_j||_2 \cos \theta_j , \quad \forall j \not =y,
	\end{align}
	where $ \theta_y$  and $\theta_{j} $  denote the angles between the feature vector $ f(x) $ and the corresponding weight vectors, respectively.

	Intuitively,
    the larger the length of $ w_y $  ,
    a smaller $ \cos \theta_y $  can ensure that \eqref{normeq2} holds.
	Nevertheless, a smaller $ \cos \theta_y $  means that the learned feature  $ f(x) $  is closer to the decision boundary.
	However,  it has already been pointed out  in \cite{moosavi2016} that the main reason why neural networks are vulnerable to attacks
    is that the learned features are too close to the decision boundary.
    Therefore,  in order to guarantee that all features are not close to the decision boundary,
    we do not want to see any longer weight vectors.
    The most ideal situation is that all vectors have the same length.
    Thus, we set
     \begin{align}
     \label{Assump}
              ||w_i||_2 = s , \quad \forall 1 \leq i \leq K,
     \end{align}
     where $ s $ is a positive hyperparameter.
     Consequently, the label of a sample $ x $  is determined by
	\begin{align*}
		\mathop{\arg \max} \limits_{1\leq i \leq K} w_i^T f(x) := \mathop{\arg \min} \limits_{1\leq i \leq K}\|f(x) - w_i\|_2^2,
	\end{align*}
	where we have used the condition \eqref{Assump} in the equality.

	Then, we come to  the first optimization objective
	\begin{align}
		\label{intra-class}
		\min_{ f} \mathbb{E}_{(x, y)} \|f(x) - w_y\|_2^2,
	\end{align}
    which encourages the intra-class compactness of the  feature representations.

   On the other hand,
   it is obvious that the greater the distance between the class centers,  the harder it  is to confuse the classification results.
   Therefore, in order to further improve the robustness,
   we expect the distance between any two class centers to be as large as possible.
   To this end, we formulate the second optimization objective as follows
   	\begin{align}
		\label{inter-class}
		\max_W \min_{1\leq i<j \leq K} \|w_i - w_j\|_2^2,
	\end{align}
   which encourages the inter-class diversity of the feature representations.

   So far,  from an intuitive perspective,
   we have derived the optimization objectives \eqref{intra-class} and \eqref{inter-class} that are subject to the constraint \eqref{Assump},
   which  are same as those in \cite{pang2018,pang2020}.

\subsection{The construction of orthogonal robust classifier}
\label{sec:construction}
According to the arguments in Section 3.1,
the weights of a robust classifier should satisfy \eqref{Assump} and \eqref{inter-class}.
	
It has been proved in  \cite{pang2018} that under the constraint  \eqref{Assump},
the maximal value of \eqref{inter-class} is  $ 2K s^2/(K-1) $,
where $ K $ is the number of classes.
Moreover,
an algorithm named as GenerateOptMeans was used in  \cite{pang2018,pang2020}
to explicitly construct a specific \emph{upper triangular} matrix $W $ to satisfy \eqref{Assump} and \eqref{inter-class}.
After then, the classification layer is frozen  and only the encoder part is trained to satisfy the intra-class loss \eqref{intra-class}.

However, we point out here that the upper triangular structure adopted in \cite{pang2018,pang2020} may cause some problems.
Consider a simple neural network with only one linear encoder $ f(x) =V^T x $.
Let $ W \in \mathbb{R}^{P \times K} $ denote the corresponding classification layer.
When  $ W $  is an upper triangular matrix and the feature dimension $ P $ is larger than the category number $ K $ (a very common case in practice),
the classifier can be simplified as
\begin{align*}
		c(x) = W^T V^T x:=\widetilde{W}^T \widetilde{V}^T x,
\end{align*}
where $\widetilde{W}$ is a submarix constituted by the first $ K $ rows of $ W $,
and $\widetilde{V}$ only includes the first $ K $  columns of $ V $.
Thus, there are  $ (P-K) $ features \emph{ unused} for classification at all,
which can lead to underfitting  in some situations (see Section \ref{experiment:redundancy} for related experiments).
We call the above phenomenon as structural redundancy.
	
	\begin{algorithm}[htb]
   		\caption{The Construction of a Dense Orthogonal Classifier}
        \label{alg:constrution}
		{\bf Input:}  The number of classes $ K $,  the length $ s $ and the feature dimension $P$.
		\begin{algorithmic}
			\State Set $ T = \log_2 P $ and $ M^{(0)} = 2^{-T/2} s $.
			\For{$t=1,2, \cdots, T$}
			\State Apply the formula \eqref{construction} to recursively construct a series of $ M^{(t)} $.
			\EndFor
			\State Choose the first $ K $ columns of $ M^{(T)}$  to build the weight matrix $ W $.
		\end{algorithmic}
		{\bf Output:} The weight matrix $ W $ for classifier.
	\end{algorithm}

\begin{rmrk}
	Though Algorithm \ref{alg:constrution} takes the number of classes $K$, the length $s$ and the feature dimension $P$ as inputs,
	$K$ and $P$ are directly determined by the learning task and the neural network, respectively.
	Only the vector length $s$  is left as a hyperparameter to be given.
\end{rmrk}

In order to overcome the problem of structural redundancy,
we next propose a step by step procedure to construct a \textit{fully dense} weight matrix $ W $
that \textit{approximately} satisfies the constraint \eqref{inter-class}.
For simplicity,  we assume that the feature dimension satisfies $ P = 2^T $, where $ T $ is a given integer.

Set $ M^{(0)}=2^{-T/2} s $.	For $ t = 1, \cdots, T $, we recursively define
	\begin{align}
		\label{construction}
		M^{(t)} =
		\left [
		\begin{array}{cc}
			M^{(t-1)} &\; -M^{(t-1)} \\
			M^{(t-1)} &\;  M^{(t-1)}
		\end{array}
		\right ].
	\end{align}
 In view of the hierarchical block structure of  $ M^{(t)} $,
it is easy to verify that
 \begin{align}	
		M^{(t)} = 2^{-T/2} s  \bigg[m_{i,j}\bigg]_{2^t \times 2^t},
	\end{align}
where $ |m_{i,j}|=1 $,
which means that $ M^{(t)} $ is a fully dense matrix with all entries of the same magnitude.

Furthermore, by an inductive argument, we have
	\begin{align*}
		( M^{(t)})^T M^{(t)}  = 2^{t-T}  s^2 I,  \quad \forall 1 \leq t \leq T,
	\end{align*}
which means that the column vectors of $ M^{(t)}$ are orthogonal to each other.
	
Finally,  we choose the first $ K $ columns of $ M^{(T)} $ to formulate the  desired weight matrix $ W $ for the classifier.

\begin{rmrk}
    Our approach has several advantages.
    First, the recursive construction process \eqref{construction} is simple and easy to implement.
    Second, because all entries of the constructed weight matrix are non-zero ($ |m_{i,j}|=1 $),
    there is no problem of structural redundancy.
    Third, the magnitude of all entries is the same, which forces the encoder to learn all feature representations unbiased.
\end{rmrk}

\begin{rmrk}
        It follows from orthogonality that  $ \|w_i - w_j\|_2^2 = 2 (s^2 - w_i^Tw_j)= 2s^2 $ for all  $ 1\leq i < j \leq K $.
		The value $ 2s^2 $ is smaller than the optimal value  $ 2K s^2/(K-1) $ of  \eqref{inter-class}.
        Nonetheless,  for a large $ K $, which is common in practice, such difference is negligible.
		Thus, the proposed weight matrix can be regarded as an approximate solution of the problem \eqref{inter-class}.
\end{rmrk}
	
\begin{figure}[!htb]
		\centering
		\begin{minipage}{0.45\textwidth}
			\centering
			\scalebox{0.8}{\includegraphics{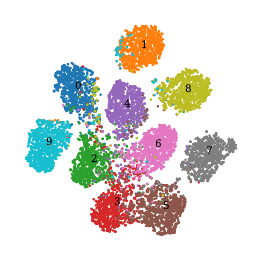}}
		\end{minipage}
		\begin{minipage}{0.45\textwidth}
			\centering
			\scalebox{0.8}{\includegraphics{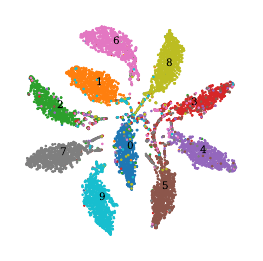}}
		\end{minipage}
		\caption{t-SNE \cite{maaten2008} visualization of the feature representations on CIFAR-10, where the index numbers indicate the corresponding classes.
			\textbf{Left}:   Feature distribution obtained by a normal training of the entire network;
			\textbf{Right}: Feature distribution learned by using the orthogonal classifier constructed by Algorithm \ref{alg:constrution}.}
		\label{features}
	\end{figure}

After constructing the desired classifier,	
one just need to fix the classifier and train the encoder part by minimizing the intra-class loss \eqref{intra-class}.
In this way, a standard training on clean data is sufficient to learn discriminative features (see Figure \ref{features}),
thereby ensuring the accuracy and certain robustness of the model.

\subsection{Using adversarial samples to further improve robustness}
In the previous section,
we only use clean data for learning.
In order to further improve the robustness,
we can turn to adversarial samples  \cite{madry2018,zhang2019b} and introduce  a special worst-case loss.

More specifically, for any clean sample $x$, let $x'$ denote the corresponding adversarial sample.
We expect that the feature representation of any adversarial sample should still satisfy
the intra-class compactness and inter-class dispersion  properties discussed in Section 3.1.
This means that  after  determining the classifier weight matrix $ W $  from Algorithm \ref{alg:constrution},
we shall solve the following optimization problem
\begin{align}
	\label{final-loss}
	\min_{f} \: \frac{1}{P}\: \mathbb{E}_{(x, y) }
      \Bigg\{ \alpha \cdot \|f(x) - w_y\|_2^2 + (1 - \alpha) \cdot \max_{\|x' - x\| \le \epsilon} \|f(x') - w_y\|_2^2 \Bigg\},
\end{align}
where  $\|\cdot\|$ denotes $\ell_{\infty}$ norm, $\epsilon$  is a perturbation budget,
and the hyperparameter $\alpha \in (0, 1]$  controls the effects of the clean samples and adversarial samples.
The second item in \eqref{final-loss} means that the worst-case feature representations should also be compact around the corresponding classification centers,
thus leading to  a stronger robustness.
Obviously, the previous optimization objective \eqref{intra-class} is a special case of $\alpha = 1$.
We explicitly multiply the coefficient $\frac{1}{P}$,
where $ P $ denotes the output feature  dimension of the network,
 to ensure consistent performance over different architectures.

\begin{rmrk}
	It is infeasible to exactly solve the second optimization term in \eqref{final-loss}.
	However, as suggested by \cite{madry2018,zhang2019b},
    we can approximately solve this problem by searching the worst-case adversarial samples on the fly,
    e.g., applying PGD attack for 10 iterations with a step size $0.25\epsilon$.
\end{rmrk}

\section{Experiments}

\subsection{Experimental setup}

\textbf{Datasets.} In this section, we evaluate the performance of our approach on three commonly used datasets,
MNIST \cite{lecun1998}, FashionMNIST \cite{fmnist2017} and CIFAR-10 \cite{krizhevsky2009}.
MNIST and FashionMNIST are collections of grayscale handwritten digits and clothes, respectively.
Both of them consist of 60000 training samples and 10000 test samples, namely 7000 $28 \times 28$ samples per class.
FashionMNIST is a more challenging task than MNIST and thus can be served as a direct drop-in replacement for it.
The CIFAR-10 dataset contains 60000 $32\times 32$ color images, which are divided into a training set of 50000 images and a test set of 10000 images.

\begin{table}[htb]
	\caption{The basic setup for adversarial attacks in $\ell_{\infty}$, $\ell_{1}$ and $\ell_{2}$ norms.}
	\label{table-settings}
	\centering
	\scalebox{.55}{
		\begin{tabular}{cccccccccccccccc}
			\toprule
			& \multicolumn{7}{c}{$\ell_{\infty}$} & & \multicolumn{2}{c}{$\ell_1$} & & \multicolumn{4}{c}{$\ell_{2}$} \\
			\cmidrule{2-8} \cmidrule{10-11} \cmidrule{13-16}
			& FGSM & PGD & PGD & PGD & PGD & DeepFool& AutoAttack & & PGD & SLIDE &  & PGD & PGD & C\&W & AutoAttack \\
			\midrule
			number of iterations & - & 10 & 20 & 40 & 50 & 50  & - & &50 & 50 & & 50& 100 & 1000 & - \\
			step size & - & 0.25 & 0.1 & 0.033333 & 0.033333 & 0.02& - & &0.05 & 0.05 & & 0.1 & 0.05 & 0.01 & -  \\		
			\bottomrule
	\end{tabular}}
\end{table}

\textbf{Attacks.}
To reliably evaluate the adversarial robustness of defense methods,
several benchmark adversarial attacks including
FGSM \cite{goodfellow2015}, PGD \cite{madry2018}, DeepFool \cite{moosavi2016}, C\&W \cite{carlini2017},
SLIDE \cite{tramer2019} and AutoAttack \cite{croce2020} are adopted.
Except that AutoAttack is due to the source code from \cite{croce2020},
all implementations of them are provided by FoolBox \cite{foolbox2017}.
The major settings of these attacks are listed in Table \ref{table-settings},
wherein step size denotes the relative step size of PGD and SLIDE,
the learning rate of C\&W, and the overshoot of DeepFool, respectively.
For brevity, refer to PGD20 as the shorthand of PGD with 20 iterations.

\textbf{Baseline defenses.} We select 5 benchmark defenses,
FGSM-AT \cite{goodfellow2015}, PGD-AT \cite{madry2018}, ALP \cite{kannan2018}, TRADES \cite{zhang2019b} and MMC \cite{pang2020},
for comparison.
We implement these defense methods following the default settings of original papers.
In particular, we train FGSM-AT and ALP using the same setup as PGD-AT.
All methods performed on FashionMNIST follow the same training settings for MNIST.
When adversarial samples are required in the procedure of training,
FGSM-AT crafts perturbations on the fly by FGSM attack,
while others utilize PGD10 attack within $\epsilon=8/255$ on CIFAR-10
and PGD40 attack within $\epsilon=0.3$ on MNIST and FashionMNIST.
Note that although we mainly focus on $\ell_{\infty}$ distance in this article,
we also report adversarial robustness over $\ell_1, \ell_2, \ell_{\infty}$ norms for completeness.

\textbf{Architectures.}
Following MMC \cite{pang2020},
we verify the effectiveness of our approach on CIFAR-10 using ResNet32,
of which the dimension is $P=64$ and accordingly $T=\log_2 P = 6$.
When training on CIFAR-10, we use the augmentation popularized by Residual Networks:
4 pixels are reflection padded on each side, and a $32 \times 32$ crop is randomly sampled from the padded image or its horizontal flip.
According to the suggestion in \cite{carlini2017,zhang2019b},
we adopt a small CNN for evaluation on MNIST and FashionMNIST,
which comprises four convolutional layers followed by three linear layers.
In order to satisfy the condition $P=2^T$, we adjust its original dimension $P=200$ to $P=256$,
and accordingly $T=\log_2 P = 8$.
In addition, we employ PReLU\cite{he2015} as the activation function instead of widely used ReLU \cite{nair2010},
because the latter only outputs positive signals, leading to an unbalanced distribution of features.

\textbf{Hyperparameters.}
As suggested in \cite{pang2018}, the length of weight vectors is set as $s=10$.
By searching $\alpha$ with a step size of $0.05$,
we find the model is optimal as long as $\alpha$ is roughly between 0.1 and 0.4.
We train ResNet32 for 180 epochs
using the SGD optimizer with an initial learning rate 0.1 which is decayed by a factor 10 at epoch 100, 150 and 175.
The MNIST and FashionMNIST datasets are simpler than CIFAR-10,
so it is sufficient to run for 80 epochs through the SGD optimizer
with an initial learning rate 0.1 which is decayed by a factor 10 at epoch 35, 60 and 75.

\subsection{Comparison with GenerateOptMeans}
\label{experiment:redundancy}
In this section, we compare Algorithm \ref{alg:constrution} with GenerateOptMeans  \cite{pang2020} in three aspects:
the speed of weight matrix construction,  structural redundancy issue and robustness performance.

\begin{figure}[htb]
	\centering
	\scalebox{.8}{\includegraphics{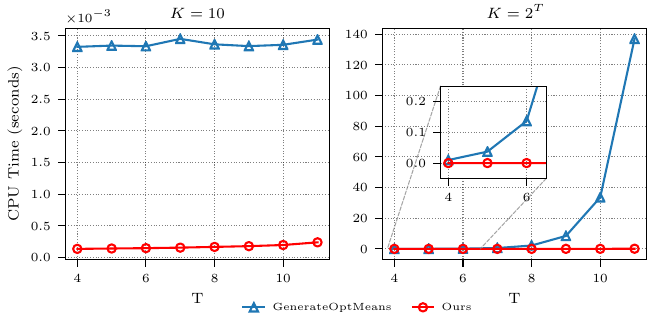}}
	\caption{
	Comparison of the  speed of constructing a weight matrix $W \in \mathbb{R}^{2^T \times K}$ under various $ T $,
    where $T=\log_2 P$ and $ P $ is the output feature dimension of the encoder.
	\textbf{Left:}  Fixed $K=10$;
	\textbf{Right:} $ K $ is increasing with $ T $.
	}
	\label{fig-genTime}
\end{figure}

\textbf{Construction speed.} We first compare the CPU Times required by the two construction algorithms.
All CPU Times (in seconds) are averaged over a total of 10 trials.
It can be seen from Figure \ref{fig-genTime} that when the number of classes is fixed at $K=10$,
although both algorithms only require milliseconds, our construction algorithm is still much faster than GenerateOptMeans \cite{pang2020}.
Moreover, in an artificial case of $ K=2^T $,
we find that the speed advantage of Algorithm \ref{alg:constrution} becomes more apparent as the number of categories increases.

\begin{figure}[htb]
	\centering
	\scalebox{1.}{\includegraphics{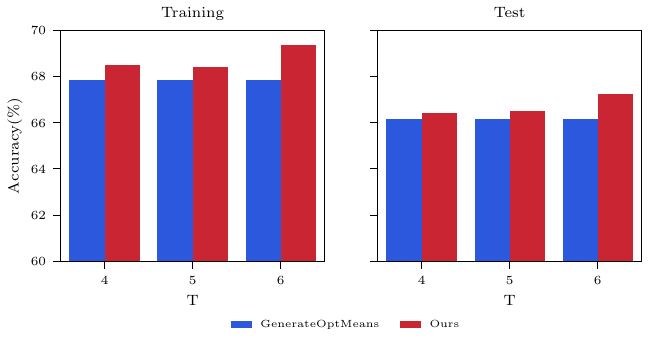}}
	\caption{Compared accuracy on a toy example with increasing feature dimensions.
                 \textbf{Left:} Training accuracy; \textbf{Right:} Test accuracy.
                 For GenerateOptMeans, the structural redundancy of classifier weights leads to underfitting on natural accuracy.
              }
	\label{STRed}
\end{figure}

\textbf{Structural redundancy issue.} We perform a toy experiment on MNIST to illustrate this problem.
We take the first 2000 images from MNIST for training and use the entire test set for evaluation.
A linear fully-connected network with only one hidden layer is adopted here.
The input of the encoder part is a 16-dimensional vector (that is, the image is resized to $4 \times 4 $ and then flattened into a vector)
and the output dimension of the encoder is set to $ P=2^T $.
We utilize Algorithm \ref{alg:constrution} and the GenerateOptMeans algorithm \cite{pang2018}
to calculate the classifier, respectively.
Then a standard training  on encoder part is conducted for 10 epochs with a learning rate 0.1.
The training and test accuracies averaged over 5 independent experiments are depicted  in Figure \ref{STRed}.

Under a normal situation,
as the dimensionality of feature representation increases,
the representation capability of the encoder will be improved accordingly,
which ultimately leads to an improvement in the final classification accuracy.
However, we observe from Figure \ref{STRed} that  with the increase of  $ T $,
the GenerateOptMeans-based networks maintain an abnormally constant classification accuracy,
while our approach yields a normal and better performance.
Such phenomenon shows that the upper triangular weight matrix constructed in \cite{pang2018,pang2020}
will indeed cause the problem of underfitting.

\begin{table}[!htb]
	\center
	\caption{Comparison of classification accuracy (\%) between MMC and the proposed method on CIFAR-10.
		We directly cite the * results of MMC from RobustBench \cite{croce2020b} and others from the original paper.
	}
	\label{table-redundancy}
	\scalebox{.8}{
		\begin{tabular}{ccccccc}
			\toprule
	
	&	& \multicolumn{2}{c}{$\ell_{\infty}$, $\epsilon=8/255$}	&& \multicolumn{1}{c}{$\ell_{\infty}$, $\epsilon=16/255$} & \\
			\cmidrule{3-4}	\cmidrule{6-6}
	& Clean & PGD50 & AutoAttack && PGD50  &\\
			\midrule
MMC \cite{pang2020}   & $80.89^*$  & \underline{55.00} & $\underline{43.48}^*$ &&\underline{27.70} & \\
Ours ($\alpha=1$) & \textbf{92.62}  & \textbf{78.85} &0.00 && \textbf{61.87} & \\
Ours ($\alpha=0.2$)  & \underline{81.21} &46.39  &\textbf{43.64} &&24.12 & \\
\bottomrule
\end{tabular}}
\end{table}

\textbf{Robustness.} Finally, we use ResNet32 to compare the  accuracy and robustness of our approach with MMC \cite{pang2020} on the CIFAR-10 dataset.
We take bold type to indicate the best result, and underline type to indicate the second best result.
As shown in Table \ref{table-redundancy},
our approach of $\alpha=1$ significantly surpasses MMC on natural accuracy and under PGD attack.
Moreover, our approach of $\alpha=0.2$ still outperforms MMC
when exposed to the adversarial samples crafted by AutoAttack, the most powerful attack by far.

\subsection{Robustness evaluation}

In this section, we compare our approach with several baseline defenses,
including FGSM-AT \cite{goodfellow2015}, PGD-AT \cite{madry2018}, ALP \cite{kannan2018} and TRADES \cite{zhang2019b},
on CIFAR-10 and FashionMNIST datasets.
Recall that all adversarial samples are crafted in $\ell_{\infty}$ distance,
but for more reliable evaluation, we also provide the robustness in other norms.

\textbf{CIFAR-10.} We first report the results under widely used perturbations $8/255$ and $16/255$ in $\ell_{\infty}$ norm.
Table \ref{table-cifar10-linf} shows a consistent improvement on both natural accuracy and adversarial robustness.
Even under AutoAttack, one of the most aggressive attacks,
our method of $\alpha=0.15$ can still outperform PGD-AT by nearly 1.1\% and 1.4\% within $\epsilon=8/255$ and $\epsilon=16/255$, respectively.
TRADES achieves the strongest robustness within $\epsilon=16/255$,
whereas it comes at the expense of natural accuracy.
It manifests that our approach can make a better trade-off, enhancing the robustness while preserving high natural accuracy.

\begin{table}[!htb]
	\center
	\caption{Comparison of classification accuracy (\%) on CIFAR-10
	under $\ell_{\infty}$ perturbations within $\epsilon=8/255$ or $\epsilon=16/255$.	
	}
	\label{table-cifar10-linf}
	\scalebox{.7}{
		\begin{tabular}{cccccccccc}
			\toprule
		&	&	& \multicolumn{3}{c}{$\ell_{\infty}$, $\epsilon=8/255$}	& 		& \multicolumn{3}{c}{$\ell_{\infty}$, $\epsilon=16/255$}	\\
			\cmidrule{4-6}	\cmidrule{8-10}
	& Clean & & PGD20 & DeepFool& AutoAttack & & PGD20 & DeepFool & AutoAttack\\
			\midrule
Standard Training                    &\textbf{93.27}  & &0.00 &0.02&0.00 &	&0.00 &0.00 &0.00\\
FGSM-AT \cite{goodfellow2015}		 &88.23  & &0.01 &0.17 &0.00 &  &0.00 &0.00 &0.00 \\
PGD-AT \cite{madry2018}              &79.42  & &48.44&48.70&42.99&	&19.28&25.50&10.92\\
ALP     \cite{kannan2018} 			 &80.70  & &48.27&\underline{49.54}&\underline{43.23}&	&19.70&\underline{27.22}&12.27\\		
TRADES ($1/\lambda=1$) \cite{zhang2019b}&82.22  & &39.93&41.83&34.62&  &10.70&17.29&6.15\\
TRADES ($1/\lambda=6$) \cite{zhang2019b}&74.04  & &45.90&44.73&40.78&	&21.06&23.80&\textbf{14.49}\\		
\midrule
Ours ($\alpha=1$)		        	 &\underline{92.62}  & &\textbf{78.68}&32.22&0.00 &	&\textbf{61.36}&12.00&0.00\\
Ours ($\alpha=0.15$)	             &80.43  & &\underline{48.83}&\textbf{51.31}&\textbf{44.15}&	&\underline{27.50}&\textbf{32.24}&\underline{12.31}\\
\bottomrule
\end{tabular}}
\end{table}

\begin{table}[!htb]
	\center
	\caption{Comparison of classification accuracy (\%) on CIFAR-10
	under $\ell_1, \ell_2$ perturbations within $\epsilon=12$ and $\epsilon=0.5$, respectively.	
	}
	\label{table-cifar10-l12}
	\scalebox{.7}{
		\begin{tabular}{ccccccccc}
			\toprule
		&	&	& \multicolumn{2}{c}{$\ell_{1}$, $\epsilon=12$}	& 		& \multicolumn{3}{c}{$\ell_{2}$, $\epsilon=0.5$}	\\
			\cmidrule{4-5}	\cmidrule{7-9}
	& Clean & & PGD50 & SLIDE & & PGD50 & C\&W & AutoAttack\\
			\midrule
Standard Training                            &\textbf{93.27} & &0.62 &0.69 &	&0.01 &0.00 &0.00 \\
FGSM-AT \cite{goodfellow2015}				 &88.23 & &2.08 &29.54& &0.01 &0.01 &0.00 \\
PGD-AT \cite{madry2018}                 &79.42 & &57.25&23.08&	&56.70&\textbf{54.67}&53.31\\
ALP 	\cite{kannan2018}					 &80.70 & &55.68&22.16&	&56.22&54.26&\underline{53.34}\\		
TRADES ($1/\lambda=1$) \cite{zhang2019b}		 &82.22 & &58.17&18.26&	&54.68&51.88&51.26\\		
TRADES ($1/\lambda=6$) \cite{zhang2019b}		 &74.04 & &54.95&25.24&	&54.29&51.32&51.10\\		
\midrule
Ours	($\alpha=1$)	                     &\underline{92.62} & &\textbf{89.83}&\textbf{84.35}&	&\textbf{89.61}&0.92 &0.00\\
Ours   ($\alpha=0.15$)	                     &80.43 & &\underline{60.41}&\underline{30.64}&	&\underline{59.88}&\underline{54.59}&\textbf{53.48}\\
\bottomrule
\end{tabular}}
\end{table}

\begin{table}[!htb]
	\center
	\caption{Comparison of classification accuracy (\%) on FashionMNIST
	over $\ell_1, \ell_2, \ell_{\infty}$ perturbations within $\epsilon=10$, $\epsilon=2$ and $\epsilon=0.3$, respectively.
	}
	\label{table-fashionmnist-l12inf}
	\scalebox{.62}{
		\begin{tabular}{ccccccccccccccc}
			\toprule
		&	&	& \multicolumn{2}{c}{$\ell_{1}$, $\epsilon=10$}	& 		& \multicolumn{3}{c}{$\ell_{2}$, $\epsilon=2$} &  & \multicolumn{4}{c}{$\ell_{\infty}$, $\epsilon=0.3$}	\\
			\cmidrule{4-5}	\cmidrule{7-9} \cmidrule{11-14}
	& Clean & & PGD50 & SLIDE & & PGD100 & C\&W & AutoAttack & & FGSM & PGD50 & DeepFool & AutoAttack\\
			\midrule
Standard Training    &91.82 &&24.50&8.27 &&0.02 &0.00 &0.00 &&2.12 &0.00 &0.00 &0.00 \\
FGSM-AT \cite{goodfellow2015}   &\underline{91.94} &&20.72&12.75&&0.18 &0.00 &0.00 &&\textbf{96.34}&0.00 &0.02 &0.00 \\
PGD-AT  \cite{madry2018}        &77.76 &&67.20&57.99&&62.19&\textbf{48.09}&0.19 &&70.55&\underline{61.97}&\textbf{64.04}&\underline{45.99}\\
ALP     \cite{kannan2018}       &83.08 &&65.74&53.65&&64.14&25.53&2.35 &&68.21&61.40&56.35&24.25\\
TRADES ($1/\lambda=1$) \cite{zhang2019b} &86.06 &&68.88&\underline{58.40}&&65.20&26.61&\underline{3.46} &&67.89&58.07&56.13&29.61\\
TRADES ($1/\lambda=6$) \cite{zhang2019b} &78.05 &&63.98&50.34&&56.30&22.15&1.65 &&62.49&56.36&54.91&34.22\\
\midrule
Ours ($\alpha=1$)    &\textbf{92.56} &&\textbf{89.81}&\textbf{87.90}&&\textbf{82.91}&0.00 &0.00 &&\underline{88.44}&34.14&0.00 &0.00 \\
Ours ($\alpha=0.3)$  &78.80 &&\underline{70.10}&51.81&&\underline{68.41}&\underline{43.55}&\textbf{4.20} &&69.71&\textbf{62.20}&\underline{63.28}&\textbf{48.35}\\
\bottomrule
\end{tabular}}
\end{table}

To further demonstrate the reliability of the robustness brought by our modified orthogonal classifier,
we investigate the performance under $\ell_1$ and $ \ell_2$ perturbations.
The results are summarized in Table \ref{table-cifar10-l12}.
Although our approach is adversarially trained under $\ell_{\infty}$ perturbations within $\epsilon=8/255$,
it still extrapolates well to other unseen attacks and norms of interest.

In addition, note that even the special case ($\alpha=1$) which only uses clean data in a standard training  can protect the networks against most typical attacks
while maintaining a  high natural accuracy comparable to the Standard Training.
Recall that it only accesses the clean data and thus requires no extra computation at all.
Therefore, it may be an acceptable \textit{temporary} defense method
when dealing with tasks that place more emphasis on classification accuracy or computational load.

\textbf{FashionMNIST.} Table \ref{table-fashionmnist-l12inf}  shows the results on the FashionMNIST dataset.
Although TRADES of $1/\lambda=1$ and ALP achieve  better natural accuracy,
they perform poorly under AutoAttack, 18\% and 24\% less than ours, respectively.
TRADES ($1/\lambda=6$) does put more emphasis on robustness,
but at the cost of 8\% natural accuracy and lower robustness in $\ell_2$ norm.
Because FGSM-AT injects robustness by the simplest one-step attack,
it performs as poorly as the Standard Training when exposed to multi-step attacks.
But interestingly, its classification accuracy under FGSM attack
is significantly higher than others, and even exceeds the best natural accuracy.
This phenomenon may stem from the fact that adversarial samples could be regarded as special augmentations.
Moreover, the FGSM attack is so simple that the adversarially trained model could fit the underlying distribution easily.
In fact, even more aggressive attacks such as PGD10 can also be used to improve the image recognition \cite{xie2020}.

\subsection{Sensitivity analysis}

\begin{table}[htb]
	\center
	\caption{Comparison of classification accuracy (\%) across different hyperparameters $s$ and $\alpha$ on CIFAR-10.}
	\label{table-ablation}
	\scalebox{.7}{
		\begin{tabular}{cccccccc}
			\toprule
		&\multicolumn{7}{c}{$ \alpha=0.15$ } \\
			\cmidrule{2-8}
$s$	&  7 & 8 & 9 &  10 & 11 & 12  & 13 \\
			\midrule
Clean & 80.110  & 80.230  & 80.490 & 80.430 & 80.540 & 80.180 & 80.610 \\
AutoAttack($\ell_{\infty}$) & 44.020 & 43.770  & 43.930	& 44.150 & 43.480 & 40.060	& 43.680 \\
AutoAttack($\ell_2$) & 52.580  & 52.620  & 53.290 & 53.480 & 53.150 & 52.620 & 52.910 \\

\midrule
\midrule
		&\multicolumn{7}{c}{$s=10$} \\
			\cmidrule{2-8}
$\alpha$	&  0.01 & 0.05 & 0.15 & 0.2 & 0.3 & 0.4  & 0.5 \\
			\midrule
Clean & 78.480  & 78.430  & 80.430 & 81.210 & 81.700 & 80.740 & 81.840 \\
AutoAttack($\ell_{\infty}$) & 43.890  & 44.410 & 44.150 & 43.640 & 41.420 & 39.370	& 36.430 \\
AutoAttack($\ell_2$) & 52.560  & 52.520  & 53.480 & 53.540 & 52.270 & 52.210 & 51.520 \\
\bottomrule
\end{tabular}}
\end{table}

\begin{figure}[htb]
	\centering
	\scalebox{1.}{\includegraphics{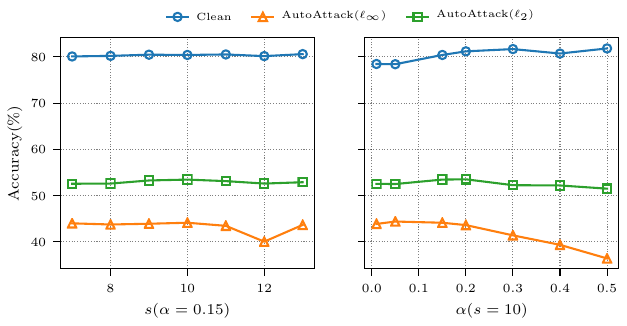}}
	\caption{Sensitivity of vector length $s$ and  loss weight $\alpha$ on CIFAR-10.
		\textbf{Left:} Classification accuracy under various $s$ with fixed $\alpha=0.15$;
		\textbf{Right:} Classification accuracy under various  $\alpha$ with fixed $s=10$.}
	\label{fig-ablation}
\end{figure}

\begin{figure}[htb]
	\centering
	\scalebox{.7}{\includegraphics{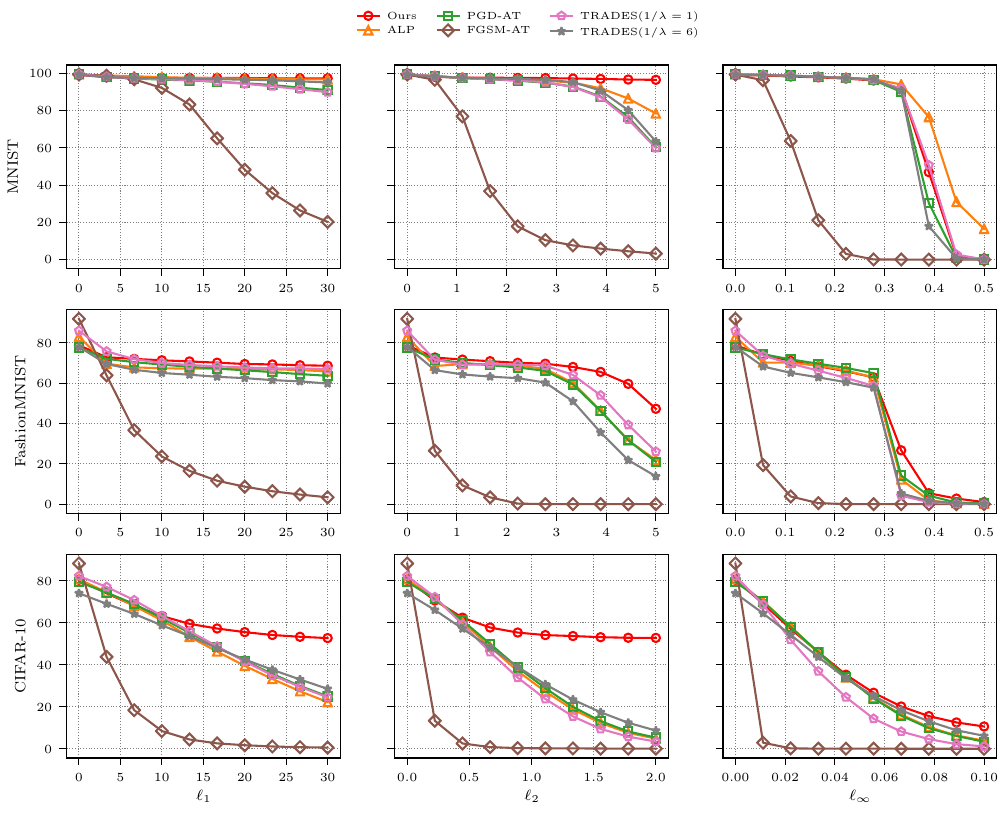}}
	\caption{Evaluation of adversarial robustness under PGD10 attack over different datasets and norms.
	Each subplot manifests the relationship regarding classification accuracy versus various perturbation budgets.
	}
	\label{fig-dea}
\end{figure}

\textbf{Hyperparameter analysis.} We first examine the effect of the hyperparameters $s$ and $\alpha$, respectively.
Experimental results are summarized in Table \ref{table-ablation} and Figure \ref{fig-ablation}.
As the length $s$ increases from 7 to 13,
both the natural accuracy and robustness change very little,
which means that the proposed method is stable w.r.t. the length of the weight vectors.
On the other hand, the balance coefficient $\alpha$,  which controls the effects of clean samples and adversarial samples, is a relatively crucial factor.
A larger $\alpha$ generally leads to better natural accuracy but less robustness.
Nevertheless, when it is roughly between 0.1 and 0.4,
the proposed method is able to maintain stable and extraordinary performances.

\textbf{Perturbation budgets.}
Although all methods are adversarially trained within $\epsilon = 8 / 255$,
it would be better if they could extrapolate to lager perturbations.
Figure \ref{fig-dea} depicts the relationship between classification accuracy and increasing perturbation budgets under PGD10 attack.
Except for the subplot at the upper right,
the proposed approach ($\alpha=0.3 $ on MNIST and FasionMNIST, and $\alpha=0.15 $ on CIFAR-10) consistently surpasses other defenses,
which demonstrates a better scalability of our approach.

\begin{figure}[htb]
	\centering
	\scalebox{.8}{\includegraphics{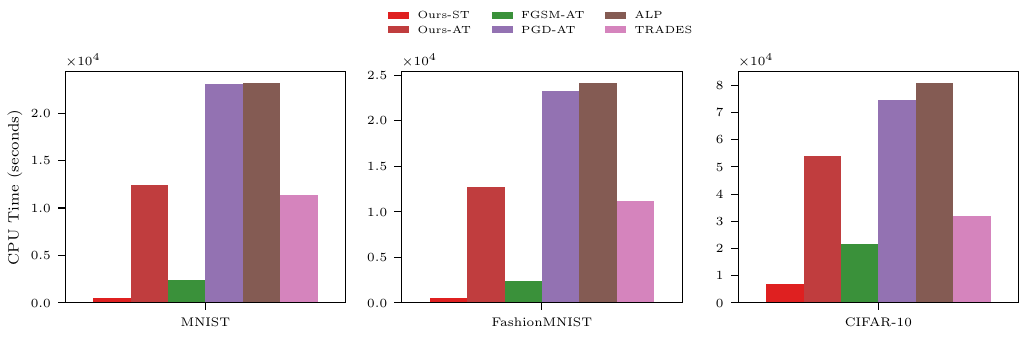}}
	\caption{Summary of computational cost on different datasets.
    Here, Ours-ST indicates a standard training using only clean data ($ \alpha =1 $),
	while Ours-AT  means using extra adversarial samples  ($ \alpha <1 $).
    We use TRADES to represent TRADES ($1/\lambda=1$) or TRADES ($1/\lambda=6$),
	as the hyperparameter $\lambda$ has little effect on CPU Time of TRADES.	
	}
	\label{fig-cputime}
\end{figure}

\subsection{Computational cost}
In this section, we list the CPU Times of different defensive approaches.
As shown in Figure \ref{fig-cputime},
the proposed approach ($\alpha=1$) only takes a few minutes for training,
while other defensive methods usually need hours at least.
Thus, our approach is a very efficient \textit{temporary} defense when only clean samples are available.

On the other hand, if a more reliable defense is required,
we have to use extra adversarial samples, which inevitably leads to more computational costs.
In this case, our approach  ($\alpha<1$) has similar time complexity to other benchmark defenses.
It is worth noting that although FGSM-AT is a fast adversarial training method,
 it can only effectively defend against a simple one-step attack.

\section{Concluding Remarks}

In this paper, by analyzing the relationship between robustness and feature distributions,
we deduced several crucial  conditions w.r.t. the intra-class compactness and inter-class dispersion, respectively.
To explicitly meet these conditions,
we then developed a simple step by step procedure to construct a fixed orthogonal classifier,
which takes only milliseconds from beginning to end.
Because all entries of the constructed matrix are non-zero, there is no problem of structural redundancy.
Thanks to the orthogonality of the weight matrix and the fact that all entries are of the same magnitude,
only using clean data in a standard training can protect networks against most typical attacks while maintaining a high natural accuracy.

To further improve the robustness, we turn to adversarial samples and introduce a special worst-case loss,
of which the hyperparameter $\alpha$ controls the balance between natural accuracy and adversarial robustness.
According to the sensitivity analysis, we found this adversarially trained model was able to maintain stable and extraordinary performances
as long as $\alpha$ was roughly between 0.1 and 0.4.

We leave a number of issues for future research.
As is the case in many adversarial training mechanisms,
our approach also suffers from the problem of the heavy computational cost
when it utilizes adversarial samples for stronger robustness.
Thus, exploring simpler and more scalable method deserves a further study in the future.
On the other hand, our approach is based on the prior assumption that
the robust category features should be scattered  and  orthogonal to each other.
Such a strong assumption will appear unreasonable when dealing with thousands of fine-grained categories.
Prior assumptions extracted from the dataset could be more appropriate and scalable.

\bigskip
\textbf{Acknowledgments}. This research is partially supported by National Natural Science Foundation of China (11771257) and Natural Science Foundation of Shandong Province (ZR2018MA008)


\begin{thebibliography}{99}

\bibitem{athalye2018}
Athalye A., Carlini N. \& Wagner D.
Obfuscated gradients give a false sense of security: circumventing defenses to adversarial examples.
In \textit{International Conference on Machine Learning (ICML)}, 2018.

\bibitem{bansal2018}
Bansal N., Chen X. \& Wang Z.
Can we gain more from orthogonality regularizations in training deep CNNs?
In \textit{Advances in Neural Information Processing Systems (NIPS)}, 2018.

\bibitem{carlini2017}
Carlini N. \& Wagner D.
Towards evaluating the robustness of neural networks.
In \textit{IEEE Symposium on Security and Privacy (SP)}, 2017.

\bibitem{carlini2019}
Carlini N., Athalye A., Papernot N., Brendel W., Rauber J., Tsipras D., Goodfellow I., Madry A. \& Kurakin A.
On evaluating adversarial robustness.
\textit{arXiv preprint arXiv:1902.06705}, 2019.

\bibitem{cohen2019}
Cohen J., Rosenfeld E \& Kolter J.
Certified adversarial robustness via randomized smoothing.
In \textit{International Conference on Machine Learning (ICML)}, 2019.

\bibitem{croce2020}
Croce F. \& Hein M.
Reliable evaluation of adversarial robustness with an ensemble of diverse parameter-free attacks.
In \textit{International Conference on Machine Learning (ICML)}, 2020.

\bibitem{croce2020b}
Croce F., Andriushchenko M., Sehwag V., Flammarion N., Chiang M., Mittal P. \& Hein M.
RobustBench: a standardized adversarial robustness benchmark.
\textit{arXiv preprint arXiv:2010.09670}, 2020.

\bibitem{goodfellow2015}
Goodfellow I., Shlens J. \&  Szegedy C.
Explaining and harnessing adversarial examples.
In \textit{International Conference on Learning Representations (ICLR)}, 2015.

\bibitem{guo2019}
Guo F., Zhao Q., Li X., Kuang X., Zhang J., Han Y. \& Tan Y.
Detecting adversarial examples via prediction difference for deep neural networks.
\textit{Information Sciences}, vol. 501, pp. 182-192, 2019.

\bibitem{gowal2019}
Gowal S., Dvijotham K., Stanforth R., Bunel R., Qin C., Uesato J., Arandjelovic R., Mann T. \& Kohli P.
Scalable verified training for provably robust image classification.
In \textit{IEEE International Conference on Computer Vision (ICCV)}, 2019.

\bibitem{han2019}
Han K., Li Y. \& Hang J.
Adversary resistant deep neural networks via advanced feature nullification.
\textit{Knowledge-Based Systems}, vol. 179, pp. 108-116, 2019.

\bibitem{he2015}
He K., Zhang X., Ren S. \&  Jian S.
Delving deep into rectifiers: surpassing human-level performance on ImageNet classification.
In \textit{International Conference on Computer Vision (ICCV)}, 2015.

\bibitem{he2016}
He K., Zhang X., Ren S. \&  Jian S.
Deep residual learning for image recognition.
In \textit{IEEE Conference on Computer Vision and Pattern Recognition (CVPR)}, 2016.

\bibitem{huang2018}
Huang L., Liu X., Lang B., Yu A. \& Li B.
Orthogonal weight normalization: solution to optimization over multiple
dependent stiefel manifolds in deep neural networks.
In \textit{Conference on Artificial Intelligence (AAAI)}, 2018.

\bibitem{krizhevsky2009}
Krizhevsky A. \&  Hinton G. E.
Learning multiple layers of features from tiny images.
\textit{Techinical report, Citeseer}, 2009.

\bibitem{kannan2018}
Kannan H., Kurakin A. \& Goodfellow I.
Adversarial logit pairing.
\textit{arXiv preprint arXiv:1803.06373}, 2018.

\bibitem{lecun1998}
LeCun Y., Bottou L., Bengio Y. \&  Haffner P.
Gradient-based learning applied to document recognition.
In \textit{Proceedings of the IEEE}, 1998.

\bibitem{moosavi2016}
Moosavi-Dezfooli S., Fawzi A. \&  Frossard P.
DeepFool: a simple and accurate method to fool deep neural networks.
In \textit{IEEE Conference on Computer Vision and Pattern Recognition (CVPR)}, 2016.

\bibitem{madry2018}
Madry A., Makelov A., Schmidt L. \& Tsipras D.
Towards deep learning models resistant to adversarial attacks.
In \textit{ International Conference on Learning Representations (ICLR)}, 2018.

\bibitem{mustafa2019}
Mustafa  A., Khan  S., Hayat M., Goecke R., Shen J. \&  Shao  L.
Adversarial defense by restricting the hidden space of deep neural networks.
In \textit{International Conference on Computer Vision (ICCV)}, 2019.

\bibitem{nair2010}
Nair V. \& Hinton G. E.
Rectified linear units improve restricted boltzmann machines.
In \textit{International Conference on Machine Learning (ICML)}, 2010.

\bibitem{naveiro2019}
Naveiro R., Redondo A., R\'{i}os Insua D., Ruggeri F.
Adversarial classification: an adversarial risk analysis approach.
\textit{International Journal of Approximate Reasoning}, vol. 113, pp. 133-148, 2019.

\bibitem{papernot2016b}
Papernot N., Mcdaniel P., Wu X., Jha S. \& Swami  A.
Distillation as a defense to adversarial perturbations against deep neural networks.
In \textit{IEEE European Symposium on Security and Privacy}, 2016.

\bibitem{pang2018}
Pang T., Du C. \& Zhu J.
Max-mahalanobis linear discriminant analysis networks.
In \textit{International Conference on Machine Learning (ICML)}, 2018.

\bibitem{pang2020}
Pang T., Xu K., Dong Y., Du C., Chen N. \& Zhu J.
Rethinking softmax cross-entropy loss for adversarial robustness.
In \textit{International Conference On Learning Representations (ICLR)}, 2020.

\bibitem{foolbox2017}
Rauber J., Brendel W. \&  Bethge M.
Foolbox v0.8.0: A Python toolbox to benchmark the robustness of machine learning models.
\textit{arXiv preprint arXiv:1707.04131}, 2017.

\bibitem{szegedy2013}
Szegedy C., Zaremba W., Sutskever I., Bruna J., Erhan D.  Goodfellow I. \& Fergus R.
Intriguing properties of neural networks.
In \textit{International Conference on Learning Representations (ICLR)}, 2014.

\bibitem{su2018}
Su D., Zhang H., Chen H.,  Yi J., Chen P. \& Gao Y.
Is robustness the cost of accuracy?-a comprehensive study on the robustness of 18 deep image classification models.
In \textit{Proceedings of the European Conference on Computer Vision (ECCV)}, 2018.

\bibitem{shafahi2019}
Shafahi A., Najibi M., Ghiasi A., Xu Z., Dickerson J., Studer C., Davis L., Taylor G. \& Goldstein T.
Adversarial training for free.
In \textit{Advances in Neural Information Processing Systems (NIPS)}, 2019.

\bibitem{tramer2019}
Tram\`{e}r F. \& Boneh D.
Adversarial training and robustness for multiple perturbations.
In \textit{Advances in Neural Information Processing Systems (NIPS)}, 2019.

\bibitem{maaten2008}
Van der Maaten L. \&  Hinton G. E.
Visualizing data using t-SNE.
\textit{Journal of Machine Learning Research}, vol. 9, pp. 2579-2605, 2008,

\bibitem{wan2018}
Wan W., Zhong Y., Li T. \& Chen J.
Rethinking feature distribution for loss functions in image classification.
In \textit{IEEE Conference on Computer Vision and Pattern Recognition (CVPR)}, 2018.

\bibitem{wang2019}
Wang Jia., Chen Y., Chakraborty R. and Yu S.
Orthogonal convolutional neural networks.
In \textit{IEEE Conference on Computer Vision and Pattern Recognition (CVPR)}, 2019.

\bibitem{wang2020}
Wang Y., Zou D., Yi J., Bailey J., Ma X. \& Gu Q.
Improving adversarial robustness requires revisiting misclassified examples.
In \textit{International Conference on Learning Representations (ICLR)}, 2020.

\bibitem{wang2021}
Wang L., Chen X., Tang R., Yue Y., Zhu Y., Zeng X. \& Wang W.
Improving adversarial robustness of deep neural networks by using semantic information.
\textit{Knowledge-Based Systems}, 2021.

\bibitem{fmnist2017}
Xiao H., Rasul K. \& Vollgraf R.
Fashion-MNIST: a novel image dataset for benchmarking machine learning algorithms.
\textit{arXiv preprint arXiv:1708.07747}, 2017.

\bibitem{xie2017}
Xie D., Xiong J. \& Pu S
All you need is beyond a good init: Exploring better
solution for training extremely deep convolutional neural networks with orthonormality and
modulation.
In \textit{IEEE Conference on Computer Vision and Pattern Recognition (CVPR)}, 2017.

\bibitem{xu2018}
Xu W., Evans D. \& Qi Y.
Feature squeezing: Detecting adversarial examples in deep neural networks.
In \textit{Network and Distributed Systems Security Symposium (NDSS)}, 2018

\bibitem{xu2020}
Xu J., Liu H., Wu D., Zhou F., Gao C. \& Jiang L.
Generating universal adversarial perturbation with ResNet.
\textit{Information Sciences}, vol. 537, pp. 302-312, 2020.

\bibitem{xie2020}
Xie C., Tan M., Gong B., Wang J., Yuille A. and \& Le Q. V.
Adversarial examples improve image recognition.
In \textit{IEEE Conference on Computer Vision and Pattern Recognition}, 2020.

\bibitem{zhang2019a}
Zhang D., Zhang T., Lu Y., Zhu Z. \& Dong B.
 You only propagate once: Accelerating adversarial training via maximal principle.
In \textit{Advances in Neural Information Processing Systems (NIPS)}, 2019.

\bibitem{zhang2019b}
Zhang H., Yu Y., Jiao J., Xing E., Ghaoui L. \& Jordan M.
Theoretically principled trade-off between robustness and accuracy.
In \textit{ International Conference on Machine Learning (ICML)}, 2019.

\end{thebibliography}
\end{document}